\documentclass[runningheads]{llncs}

\usepackage[T1]{fontenc}
\usepackage{booktabs}
\usepackage{graphicx}
\usepackage{hyperref}
\usepackage{xcolor}
\usepackage{xspace}
\usepackage{enumitem}

\newcommand{\ourtool}{\textsc{Adagio}\xspace}
\newcommand{\ourtoolcaps}{\textsc{ADAGIO}\xspace}

\begin{document}

\title{\ourtoolcaps: Interactive Experimentation with Adversarial Attack and Defense for Audio}

\titlerunning{\ourtoolcaps}

\author{
    Nilaksh Das \inst{1} 
    \and Madhuri Shanbhogue \inst{1} 
    \and Shang-Tse Chen \inst{1}
    \and Li Chen \inst{2}
    \and \\ Michael E. Kounavis \inst{2}
    \and Duen Horng Chau \inst{1}
}

\authorrunning{Das et al.}

\institute{
    Georgia Institute of Technology, Atlanta, GA, USA \\
    \email{\{nilakshdas,madhuri.shanbhogue,schen351,polo\}@gatech.edu}
    \and Intel Corporation, Hillsboro, OR, USA \\
    \email{\{li.chen,michael.e.kounavis\}@intel.com}
}

\maketitle

\begin{abstract}
Adversarial machine learning research has recently demonstrated the feasibility to confuse automatic speech recognition (ASR) models by introducing acoustically imperceptible perturbations to audio samples.
To help researchers and practitioners gain better understanding of the impact of such attacks, and to provide them with tools to help them more easily evaluate and craft strong defenses for their models,
we present \ourtool, 
the first tool designed to allow interactive experimentation with adversarial attacks and defenses on an ASR model in real time, both visually and aurally.
\ourtool incorporates AMR and MP3 audio compression techniques as defenses, which users can interactively apply to attacked audio samples.
We show that these techniques, which are based on psychoacoustic principles, effectively eliminate targeted attacks, reducing the attack success rate from 92.5\% to 0\%.
We will demonstrate \ourtool and invite the audience to try it on the Mozilla Common Voice dataset.

\end{abstract}

\section{Introduction}
\begin{figure}[t]
    \centering
    \includegraphics[width=\linewidth]{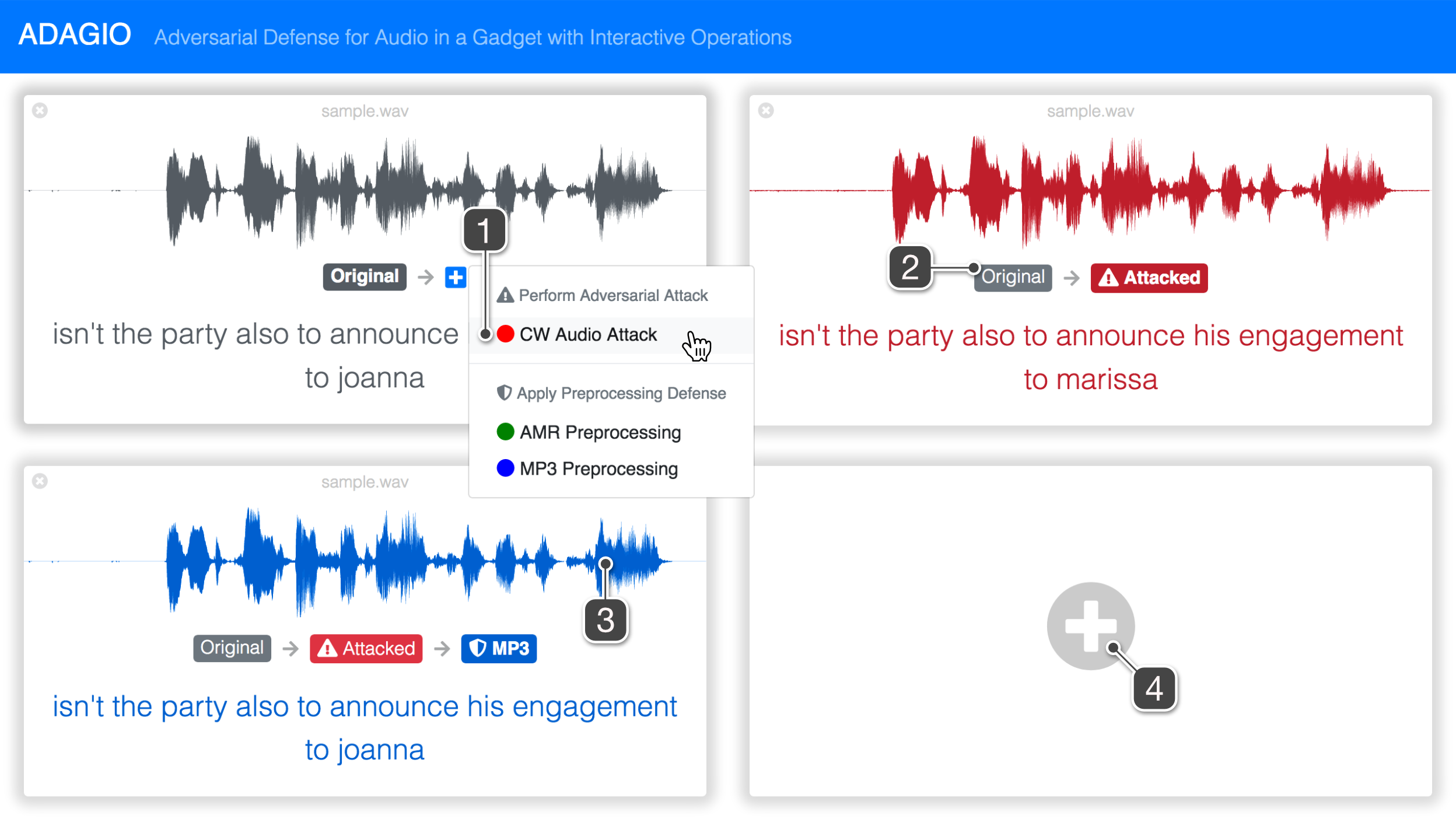}
    \caption{
    \textbf{\ourtool usage scenario}. 
    (1) 
    Jane uploads an audio file that is transcribed by DeepSpeech; then she performs an adversarial attack on the audio in real time by entering a target transcription after selecting the attack option from the dropdown menu.
    (2) 
    Jane decides to perturb the audio to change the last word of the sentence from ``joanna'' to ``marissa''; 
    she can listen to the original audio and see the transcription by clicking on the ``Original'' badge.
    (3) 
    Jane applies MP3 compression to recover the original, correct transcription from the manipulated audio;
    clicking on a waveform plays back the  audio from the selected position.
    (4) Jane can experiment with multiple audio samples by adding more cards.
    For ease of presentation, operations 1, 2 and 3 are shown as separate cards.
    }
    \label{fig:crownjewel}
\end{figure}

Deep neural networks (DNNs) are highly vulnerable to adversarial instances in the image domain \cite{goodfellow2015explaining}. Such instances are crafted by adding small imperceptible perturbations to benign instances to confuse the model into making wrong predictions.
Recent work has shown that this vulnerability extends to the audio domain \cite{carlini2018audio}, undermining the robustness of state-of-the-art models that leverage DNNs for the task of automatic speech recognition (ASR).
The attack manipulates an audio sample by carefully introducing  faint ``noise'' in the background that humans easily dismiss.
Such perturbation causes the ASR model to transcribe the manipulated audio sample as a target phrase of the attacker's choosing.
Through this research demonstration, we make two major contributions:

\medskip

\noindent \textbf{1. Interactive exploration of audio attack and defense.} 
We present \ourtool, 
the first tool designed to enable researchers and practitioners to interactively experiment with adversarial attack and defenses on an ASR model in real time (see demo: \url{https://youtu.be/0W2BKMwSfVQ}).
\ourtool incorporates AMR and MP3 audio compression techniques as defenses for
mitigating perturbations introduced by the attack.
\autoref{fig:crownjewel} presents a brief usage scenario showing how users can experiment with their own audio samples.
\ourtool stands for  \textbf{A}dversarial \textbf{D}efense for \textbf{A}udio in a \textbf{G}adget with \textbf{I}nteractive \textbf{O}perations.

\medskip

\noindent \textbf{2. Compression as an effective defense.} We demonstrate that non-adaptive adversarial perturbations are extremely fragile, and can be eliminated to a large extent by using audio processing techniques like Adaptive Multi-Rate (AMR) encoding and MP3 compression. We assume a non-adaptive threat model since an adaptive version of the attack is prohibitively slow and often does not converge.

\section{\ourtool: Experimenting with Audio Attack \& Defense}

We first provide a system overview of \ourtool, then we describe its primary building blocks and functionality. 
\ourtool consists of four major components: 
(1) an interactive UI (\autoref{fig:crownjewel});
(2) a speech recognition module; 
(3) a targeted attack generator module; and 
(4) an audio preprocessing (defense) module.
The three latter components reside on a back-end server that performs the computation.
The UI communicates the user intent with the back-end modules through a websocket messaging service, and uses HTTP to upload/download audio files for processing. 
When the messaging service receives an action to be performed from the front-end, it leverages a custom redis-based job queue to activate the correct back-end module. 
When the back-end module finishes its job, the server pings back the UI through the websocket messaging service to update the UI with the latest results.
Below, we describe the other three components in \ourtool.

\subsection{Speech Recognition}
In speech recognition, state-of-the-art systems leverage Recurrent Neural Networks (RNNs) to model audio input. 
The audio sample is broken up into frames $\{x^{(1)}, ..., x^{(T)}\}$ and fed sequentially to the RNN function $f(\cdot)$ which outputs another sequence $\{y^{(1)}, ..., y^{(T')}\}$, where each $y^{(t)}$ is a probability distribution over a set of characters.
The RNN maintains a hidden state $h^{(t)}$ which is used to characterize the sequence up until the current input $x^{(t)}$, such that, $(y^{(t)}, h^{(t)}) = f(x^{(t-1)}, h^{(t-1)})$. 
The most likely sequence based on the output probability distributions then becomes the transcription for the audio input. 
The performance of speech-to-text models is commonly measured in Word Error Rate (WER), which corresponds to the minimum number of word edits required to change the transcription to the ground truth phrase.

\ourtool uses Mozilla's implementation~\cite{mozilladeepspeech} of DeepSpeech~\cite{hannun2014deep}, a state-of-the-art speech-to-text DNN model, to transcribe the audio in real time.

\subsection{Targeted Audio Adversarial Attacks}
Given a model function $m(\cdot)$ that transcribes an audio input $x$ as a sequence of characters $y$, i.e., $m(x) = y$, the objective of the targeted adversarial attack is to introduce a perturbation $\delta$ such that the transcription is now a specific sequence of characters $y'$ of the attacker's choosing, i.e., $m(x + \delta) = y'$. 
The attack is only considered successful if there is
no error in the transcription. 

\ourtool allows users to compute adversarial samples using a state-of-the-art iterative attack \cite{carlini2018audio}. 
After uploading an audio sample to \ourtool, the user can click the attack button and enter the target transcription for the audio (see \autoref{fig:crownjewel}.1). 
The system then runs 100 iterations of the attack and updates the transcription displayed on the screen at each step to show progress of the attack.

\subsection{Compression as Defense}
In the image domain, compression techniques based on psychovisual theory have been shown to mitigate adversarial perturbations of small magnitude \cite{das2017keeping}. 
We extend that hypothesis to the audio domain and let users experiment with AMR encoding and MP3 compression on adversarially manipulated audio samples. 
Since these techniques are based on psychoacoustic principles (AMR was specially developed to encode speech), we posit that these techniques could effectively remove the adversarial components from the audio which are imperceptible to humans, but would confuse the model. 

To determine the efficacy of these compression techniques in defending the ASR model, we created targeted adversarial instances from the first 100 test samples of the Mozilla Common Voice dataset using the attack as described in \cite{carlini2018audio}.
We constructed five adversarial audio instances for every sample, each transcribing to a phrase randomly picked from the dataset,
yielding a total of 500 adversarial samples. We then preprocessed these samples before feeding it to the DeepSpeech model.
Table \ref{tab:results} shows the results from this experiment. We see that the preprocessing defenses are able to completely eliminate the targeted success rate of the attack.

\begin{table}[t]
    \centering

    \caption{Word Error Rate (WER) and the targeted attack success rate  on the DeepSpeech model (lower is better for both).
    AMR and MP3 eliminate all targeted attacks, and significantly improves WER.
    }
    \begin{tabular}{l c c c}
        \toprule
        Defense & WER (no attack) & WER (with attack) & Targeted attack success rate \\
        \midrule
        None & 0.369 & 1.287 & 92.45\% \\
        AMR & 0.488 & 0.666 & \textbf{0.00\%} \\
        MP3 & 0.400 & 0.780 & \textbf{0.00\%} \\
        \bottomrule
    \end{tabular}
    \label{tab:results}
\end{table}

\section{Conclusion}

We present \ourtool,  
an interactive tool that empowers users to
experiment  with  adversarial audio attacks  and  defenses. 
We will demonstrate and highlight \ourtool's features using a few usage scenarios
on the Mozilla Common Voice dataset, 
and invite our audience to try out \ourtool and freely experiment with their own queries.

\bibliographystyle{splncs04}
\bibliography{bibliography}

\end{document}